\documentclass[10pt,twocolumn,letterpaper]{article}

\usepackage[pagenumbers]{cvpr} 

\usepackage{svg}
\usepackage{booktabs} 
\usepackage{array}    
\usepackage{xtab}
\usepackage{longtable}
\usepackage{booktabs}
\usepackage{array}
\usepackage{amsmath}

\newcommand{\ourbench}{Gastric-X} 

\usepackage{booktabs} 
\usepackage{multirow}
\usepackage[table]{xcolor} 

\usepackage{caption}

\definecolor{lightblue}{RGB}{235,245,255} 
\definecolor{headerpurple}{RGB}{230,238,250}

\usepackage{makecell} 

\usepackage{booktabs}
\usepackage{multirow}
\usepackage{makecell}
\usepackage{threeparttable}
\usepackage{siunitx}
\newcommand{\best}[1]{\textbf{{#1}}}
\newcommand{\second}[1]{\underline{{#1}}}

\usepackage{pifont}    

\newcommand{\cmark}{\textcolor{green!60!black}{\ding{51}}} 
\newcommand{\xmark}{\textcolor{red!70!black}{\ding{55}}}   

\newcommand{\io}{Image Only}
\newcommand{\itb}{Image + Table}
\newcommand{\ibb}{Image + BBox}
\newcommand{\itbb}{Image + Table + BBox}

\definecolor{cvprblue}{rgb}{0.21,0.49,0.74}
\usepackage[pagebackref,breaklinks,colorlinks,allcolors=cvprblue]{hyperref}

\def\paperID{xxx} 
\def\confName{CVPR}
\def\confYear{2026}

\title{ Gastric-X: A Multimodal Multi-Phase Benchmark Dataset for \\Advancing Vision-Language Models in Gastric Cancer Analysis }

\author{
\renewcommand{\arraystretch}{1.12}
\begin{tabular}{ccc}
\parbox[t]{0.31\linewidth}{\centering
Sheng Lu$^{\dagger,}$\thanks{Equal contribution. $^\dagger$Correspondence: Lu Sheng and Yuanzhe Li ({ls12593@rjh.com.cn, yuanzhe@shuzhiweitech.com}).} \\
\small\mbox{Ruijin Hospital}
}
&
\parbox[t]{0.31\linewidth}{\centering
Hao Chen$^{*}$\\
\small\mbox{University of Cambridge}
}
&
\parbox[t]{0.31\linewidth}{\centering
Rui Yin\\
\small\mbox{Nanjing First Hospital}
}
\\[1.3em]
\parbox[t]{0.31\linewidth}{\centering
Juyan Ba\\
\small\mbox{Shenzhen University}
}
&
\parbox[t]{0.31\linewidth}{\centering
Yu Zhang\\
\small\mbox{Shanghai Jiao Tong University}
}
&
\parbox[t]{0.31\linewidth}{\centering
Yuanzhe Li$^{\dagger}$\\
\small\mbox{Ruijin Hospital}
}
\end{tabular}
}

\begin{document}
\maketitle

\begin{abstract}

Recent vision-language models (VLMs) have shown strong generalization and multimodal reasoning abilities in natural domains. However, their application to medical diagnosis remains limited by the lack of comprehensive and structured datasets that capture real clinical workflows. To advance the development of VLMs for clinical applications, particularly in gastric cancer, we introduce \ourbench, a large-scale multimodal benchmark for gastric cancer analysis providing 1.7K cases. Each case in \ourbench~includes paired resting and dynamic CT scans, endoscopic image, a set of structured biochemical indicators, expert-authored diagnostic notes, and bounding box annotations of tumor regions, reflecting realistic clinical conditions. We systematically examine the capability of recent VLMs on five core tasks: Visual Question Answering (VQA), report generation, cross-modal retrieval, disease classification, and lesion localization. These tasks simulate critical stages of clinical workflow, from visual understanding and reasoning to multimodal decision support.  Through this evaluation, we aim not only to assess model performance but also to probe the nature of VLM understanding: Can current VLMs meaningfully correlate biochemical signals with spatial tumor features and textual reports? 
We envision \ourbench~as a step toward aligning machine intelligence with the cognitive and evidential reasoning processes of physicians, and as a resource to inspire the development of next-generation medical VLMs. Dataset link: \textcolor{magenta}{https://huggingface.co/datasets/HaoChen2/Gastric-X}.

\end{abstract}

\section{Introduction}
\label{sec:intro}

\begin{figure*}[!ht]
    \centering
    \includegraphics[
    width=\linewidth,
    trim=0 0 0 0,        
    clip
    ]{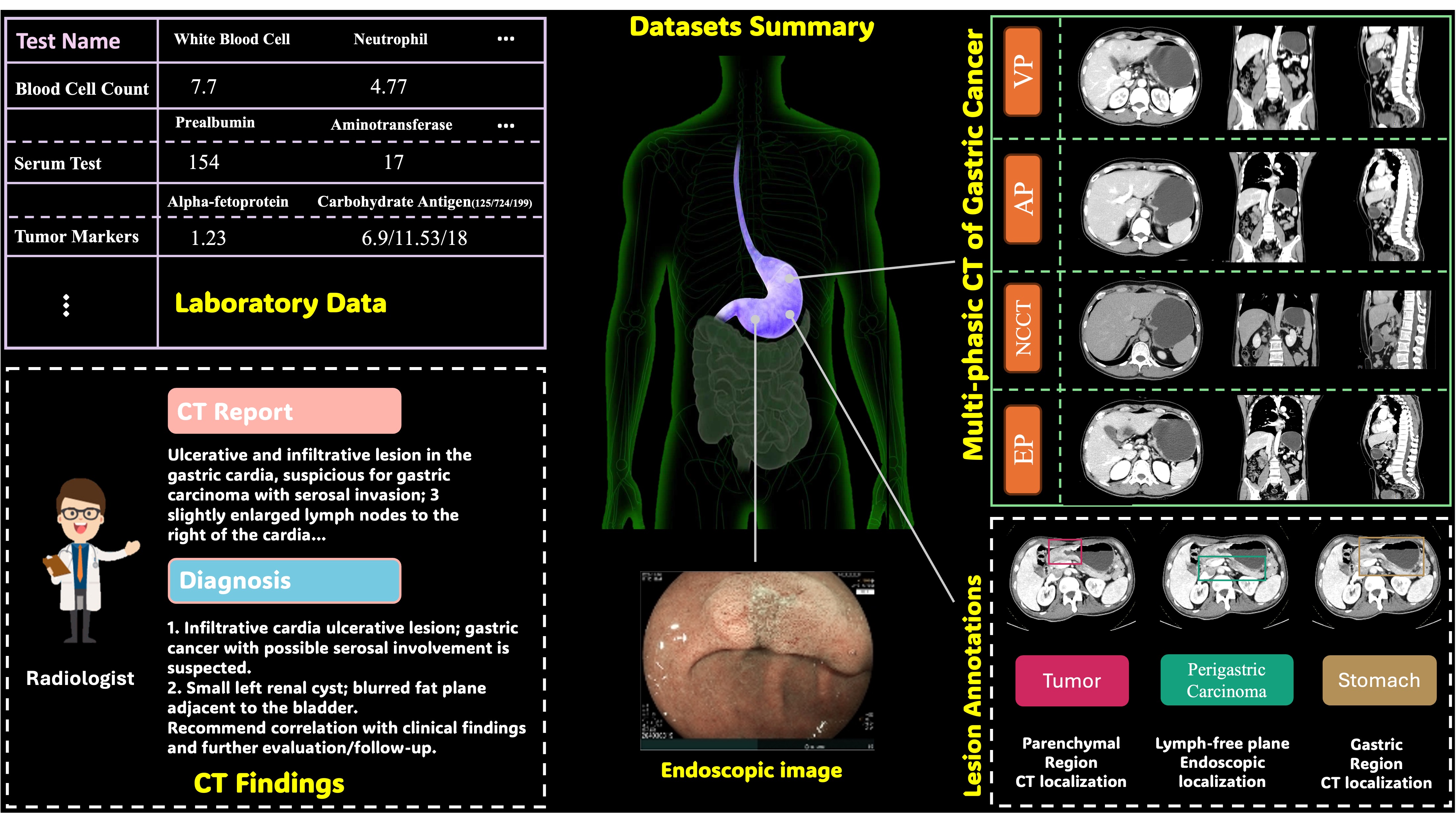}
    \caption{\textbf{Overview of the   multi-modal information  in proposed Gastric-X.} The center panel shows a schematic gastric representation alongside an endoscopic image. The left panel presents examples of structured laboratory data (e.g., blood counts, serum biochemistry, and tumor markers) and clinical textual reports (diagnostic and CT reports) that reflect real-world radiological reasoning. The right panel illustrates multi-phase 3D CT scans (non-contrast, arterial, venous, and equilibrium phases) with multi-class lesion annotations, including tumor, perigastric carcinoma, and stomach regions.}
    \label{fig:overview}
\end{figure*}


Gastric cancer remains one of the most fatal malignancies worldwide, responsible for over 769{,}000 deaths in recent global estimates \cite{chhikara2023global, bray2018globocan}. Its diagnosis requires  integrating heterogeneous, multimodal sources of evidence, including multi-phase 3D CT imaging, endoscopic assessment, laboratory tests, and patient history \cite{sexton2020gastric, kwee2007imaging, smyth2016gastric, chen2025learning, chen2026pmpbench, zhu2025medeyes}. Among these modalities,  CT plays a central role in localize lesion, clinical staging \cite{kwee2007ctstaging,liu2025m,liu2024scanext}, yet evaluations remain subject to substantial inter-observer variability \cite{kim2007diagnostic}. As the volume and complexity of patient data increase, clinicians face growing cognitive burden, and diagnostic outcomes may vary across institutions and expertise levels \cite{litjens2017survey, shen2017deeplearning, chen2024domain}. These challenges underscore an urgent need for automated systems capable of jointly perceiving and reasoning across diverse clinical data streams.


In parallel, rapid progress in vision--language models (VLMs) has reshaped our understanding of cross-modal learning in the general domain. Models such as CLIP \cite{radford2021clip}, ALIGN \cite{jia2021align}, BLIP \cite{li2022blip}, and Flamingo \cite{alayrac2022flamingo} demonstrate that large-scale alignment of images and text can unlock powerful representations capable of classification, captioning, and visual question answering. These successes have naturally inspired the development of medical VLMs \cite{muller2022joint,blankemeier2024merlinrelated3,hamamci2024foundationrelated14,lin2024ctrelate27,zhu2025pathology}. 
Yet, the datasets that currently underpin medical VLM research paint a very different picture from real-world clinical practice. Most available benchmarks focus on single-modality 2D radiographs paired with free-text reports, such as MIMIC-CXR \cite{johnson2019mimic}, CheXpert \cite{irvin2019chexpert}, and PadChest \cite{bustos2020padchest}. A few recent efforts have begun to explore volumetric data, e.g., MedVL-CT69K~\cite{shui2025largeMedVL-CT69K} and 3D-RAD~\cite{gai20253d}, but these datasets still primarily emphasize image–report matching and rarely incorporate the richer set of modalities or multi-phase volumes (except MedVL-CT69K) that are indispensable for cancer diagnosis.

This leaves a fundamental gap: \textit{clinical decision-making, especially in oncology, is inherently multimodal.} Radiologists and oncologists must integrate multi-phase 3D imaging, structured laboratory indicators, lesion localization, and clinical narratives to form a coherent diagnosis. Imaging alone provides only a partial view; without datasets that capture this complexity, VLMs often rely on superficial correlations and fail to generalize to real clinical reasoning.


To address this critical gap, we introduce \textbf{Gastric-X}, a new multimodal benchmark constructed directly from real-world gastric cancer diagnostic workflows. Gastric-X integrates four key data modalities and annotations essential to clinical decision-making: (1) multi-phase 3D CT scans (non-contrast, arterial, venous, and equilibrium phases) capturing tumor characteristics across enhancement stages \cite{kwee2007ctstaging}; (2) endoscopic images that visualize the tumor; (3) structured patient profiles and biochemical laboratory measurements used in gastric cancer staging \cite{smyth2016esmo}; and (4) comprehensive clinical and radiology reports, along with expert-annotated ground truths including disease stages and precise 3D bounding boxes of tumors, which follow the standard clinical workflow for gastric cancer localization. Additionally, the dataset provides curated VQA pairs that probe cross-modal understanding. As shown in Figure~\ref{fig:overview}, the right panel illustrates multi-phase CT imaging and annotated bounding box examples, the left panel presents structured laboratory indicators and clinical reports, and the center highlights the captured endoscopic image. This multimodal set mirrors the information pipeline encountered by radiologists in practice.


\noindent\textbf{Contributions.} The main contributions are:
\begin{itemize}
    \item We introduce \textbf{Gastric-X}, a multimodal gastric cancer dataset designed to reflect how clinicians actually reason, by integrating four modalities: multi-phase 3D CT scans, endoscopic images, laboratory measurements, and diagnostic reports, into one unified resource.
    
    \item To ground the dataset in real clinical semantics, we provide expert-crafted annotations, including precise 3D bounding boxes for tumors and perigastric lesions, detailed disease stages, and a curated set of VQA pairs that probe cross-modal understanding.

    \item Building on these components, we establish a comprehensive benchmark of five clinically meaningful tasks, offering a standardized platform for evaluating and advancing multimodal medical VLMs. 
\end{itemize}

\begin{table*}[t]
\centering
\caption{
\textbf{Comprehensive comparison of medical vision–language datasets.}
\textbf{Multi-phase:} scans or images captured at different stages or conditions.
\textbf{Biochemical data:} structured laboratory measurements (e.g., serology results) or Electronic Health Records (EHRs).
\textbf{Lesion label:} annotations of lesions, including masks or bounding boxes.
\textbf{Textual Modality:}  how text-based labels are provided.  
The VQA pairs column specifies whether the dataset is primarily designed for visual question answering.  
The report column indicates whether original diagnostic reports are available.
}

\vspace{-2.5mm}
\label{tab:dataset_comparison}
\renewcommand{\arraystretch}{1.1}
\setlength{\tabcolsep}{6pt}
\rowcolors{2}{headerpurple}{white}

\resizebox{\textwidth}{!}{%
\begin{tabular}{lcccccccccc}
\toprule

\multicolumn{4}{c}{\textbf{General Information}} & 
\multicolumn{3}{c}{\textbf{Imaging}} & 
\multicolumn{3}{c}{\textbf{Annotation \& Report}} \\ 
\cmidrule(lr){1-4} \cmidrule(lr){5-7} \cmidrule(lr){8-10}

\textbf{Dataset} & 
\textbf{Release} & 
\textbf{Domain} & 
\textbf{Link} & 
\textbf{Image Modalities} & 
\textbf{\#Scans / Images} & 
\textbf{Multi-phase} & 
\textbf{Biochemical Data} & 
\textbf{Lesion Label} & 
\textbf{Textual Modality} \\

\midrule

PathVQA~\cite{he2020pathvqa} & 2020 & Diverse & \href{https://github.com/KaveeshaSilva/PathVQA}{link} & HP & 5.00K Images & \xmark & \xmark & \xmark & VQA Pairs \\

PadChest~\cite{bustos2020padchest} & 2020 & Chest & \href{https://bimcv.cipf.es/bimcv-projects/padchest/}{link} & XR & 160K Images & \xmark & \xmark & \cmark & Reports \\

SLAKE~\cite{liu2021slake} & 2022 & Diverse & \href{https://www.med-vqa.com/slake/}{link} & CT, MR, XR & 642 Items & \xmark & \xmark & \cmark & VQA Pairs \\

Merlin~\cite{blankemeier2024merlinrelated3} & 2024 & Abdomen & \href{https://stanfordaimi.azurewebsites.net/datasets/60b9c7ff-877b-48ce-96c3-0194c8205c40}{link} & CT & 25.5K Scans & \xmark & \cmark & \xmark & Reports \\

MIMIC-CXR v2~\cite{johnson2024mimiccxr} & 2024 & Chest & \href{https://physionet.org/content/mimic-cxr/2.1.0/}{link} & XR & 377K Images & \xmark & \cmark & \xmark & Reports \\

CT-RATE~\cite{hamamci2024developing} & 2024 & Chest & \href{https://huggingface.co/datasets/ibrahimhamamci/CT-RATE}{link} & CT & 25.7K Scans & \xmark & \xmark & \xmark & Reports \\

GEMeX~\cite{liu2025gemex} & 2025 & Chest & \href{https://www.med-vqa.com/GEMeX/}{link} & XR & 151K Images & \xmark & \xmark & \xmark & VQA Pairs \\

MedVL-CT69K~\cite{cao2025boosting} & 2025 & Diverse & -- & CT & 272K Scans & \cmark & \xmark & \xmark & Reports \\

3D-RAD~\cite{gai20253d} & 2025 & Diverse & \href{https://huggingface.co/datasets/Tang-xiaoxiao/3D-RAD}{link} & CT & 16.1K Scans & \xmark & \xmark & \xmark & VQA Pairs \\

\midrule

\ourbench & 2025 & Gastric & -- & CT &
\begin{tabular}[c]{@{}c@{}}7.1K Scans\\1.7K  Images\end{tabular}
& \cmark & \cmark & \cmark & Reports \\

\bottomrule
\end{tabular}
}
{
\vspace{-2.5mm}
\begin{flushleft}
\small
\textit{Notes.} Diverse: includes multiple organs. 
CT: Computed Tomography; HP: Histopathology; MR: Magnetic Resonance; XR: X-ray.  
\end{flushleft}
}
\vspace{-4mm}
\end{table*}

\section{Related Work}
\label{sec:related}

\subsection{Vision-Language Models}
Vision–Language Models (VLMs) learn joint image–text representations from large-scale data, enabling strong zero-shot generalization~\cite{radford2021learningvlm1,jia2021scalingvlm2,yao2021filipvlm3} beyond traditional supervised learning~\cite{girshick2015fastvlm4}. 
The  CLIP model~\cite{radford2021clip} established contrastive learning as the foundation for aligning visual and linguistic spaces. 
Building on this, later works introduced multi-objective training~\cite{yu2022cocavlm5,singh2022flavavlm6} and unified architectures~\cite{tschannen2022imagevlm7,jang2023unifyingvlm8}, extending VLM capabilities to dense prediction tasks such as object detection~\cite{yao2022detclipvlm9,li2022groundedvlm10}. 
For downstream adaptation, lightweight techniques such as prompt tuning~\cite{zhou2022learningvlm11,zhou2022conditionalvlm12} and feature adaptation~\cite{gao2024clipvlm13} efficiently tailor pretrained models to new tasks, enabling VLMs to serve as versatile foundations for multimodal reasoning.

\subsection{VLMs in Medical Diagnosis}
Inspired by the success of general-domain VLMs, medical adaptations have sought to extend them to support clinical diagnosis.
Early studies primarily aligned 2D X-ray images with radiology reports through contrastive objectives~\cite{chen2022alignrelated8,tiu2022expertrelated36,cheng2023priorrelated10,zhou2023advancingre46}. 
Subsequent work refined these representations via fine-grained region–text correspondence~\cite{huang2021gloria,muller2022joint,wang2022multi} and integration of structured medical knowledge~\cite{li2023dynamic23,liu2024bootstrapping29,wu2023medklip42,zhang2023knowledge44,zhang2020radiology45}.  
However, most efforts remain limited to 2D imagery, with only recent attempts exploring 3D visual–language learning from computed tomography  (CT) volumes~\cite{cao2024bootstrappingrelated3,blankemeier2024merlinrelated3,hamamci2024foundationrelated14,lin2024ctrelate27}. A further challenge lies in the reliance on single-volume inputs, while real-world diagnosis often requires multiple imaging sequences. For instance, 4 MRI sequences for brain tumors or 2 (ideally 4) CT volumes for gastric cancer. These gaps highlight the need for domain-specific 3D VLM frameworks that better reflect clinical diagnostic practice.

\subsection{Multi-Modal Datasets for VLM Diagnosis}
Progress in medical VLMs is closely tied to the availability of high-quality, multi-modal datasets. 
Foundational resources such as TCIA~\cite{clark2013cancerTCIA} and TCGA~\cite{weinstein2013cancerTCGA} have provided extensive data that underpin cancer research.  
In radiology, dataset development has rapidly evolved, from early 2D efforts linking images to free-text reports, such as MIMIC-CXR~\cite{baratella2021chestMIMIC} and PadChest~\cite{bustos2020padchest}, to vision–language benchmarks with structured supervision like SLAKE~\cite{liu2021slake} and GEMeX~\cite{liu2025gemex}. 
More recently, 3D CT datasets including MedVL-CT69K~\cite{shui2025largeMedVL-CT69K} and 3D-RAD~\cite{gai20253d} have shifted attention toward volumetric reasoning and multi-modal integration. 
While these developments mark important progress toward holistic medical understanding, multi-modal datasets dedicated to cancer diagnosis such as gastric cancer remain scarce, limiting the advancement of specialized 3D VLMs in this critical domain.





\begin{figure*}[t]
    \centering
    \includegraphics[width=1.0\linewidth]{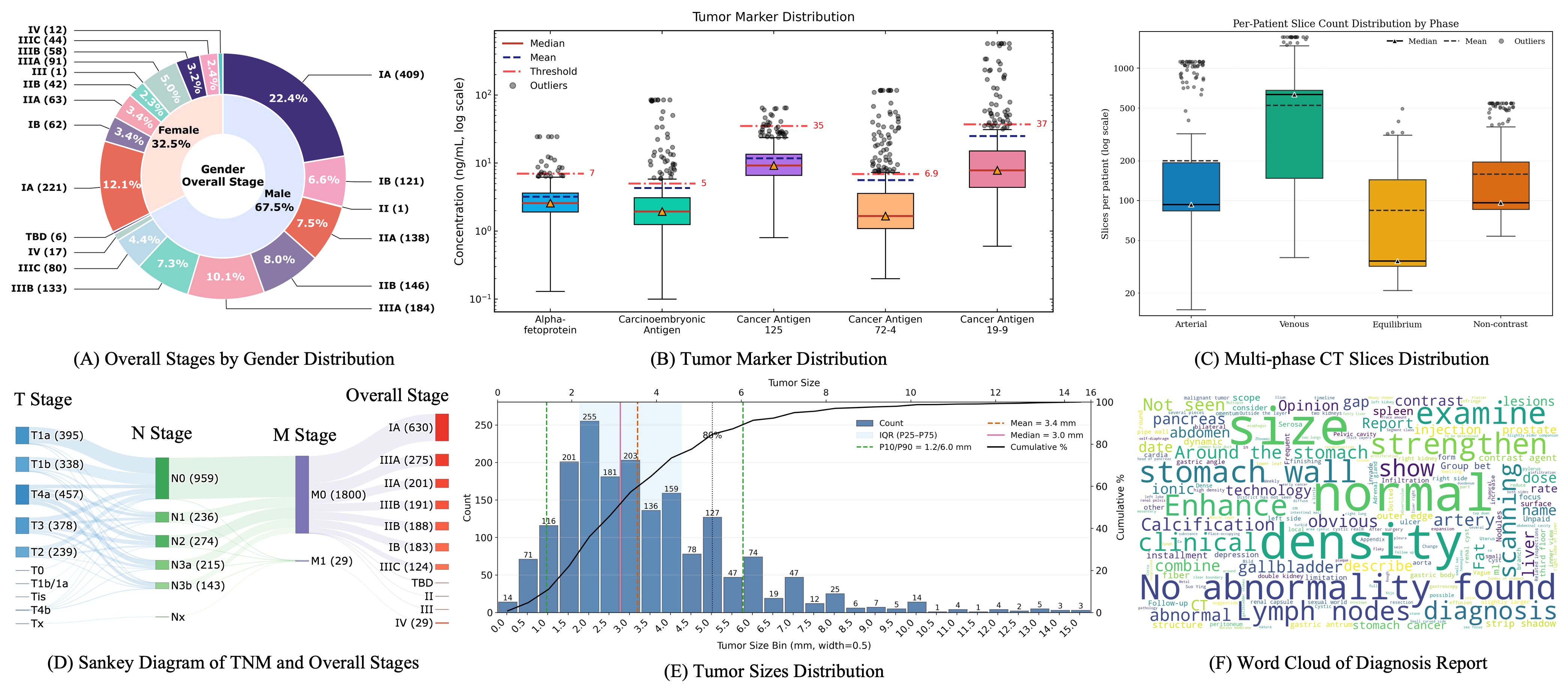}
\caption{\textbf{\ourbench~Overview.} 
(A) Overall stage distribution by gender (67.5\% male, 32.5\% female). 
(B) Distribution of 5 tumor markers on a logarithmic scale. 
(C) Overall distribution of CT slice lengths across 4 different phases. 
(D) Sankey diagram illustrating the transitions between T, N, and M stages and  corresponding overall stages. 
(E) Histogram of tumor sizes with a cumulative percentage curve. 
(F) Word cloud summarizing the most frequent terms from radiology reports. Zoom in for better visualization. }
    \label{fig:stat}
\end{figure*}

\section{Gastric-X: Design and Construction}
\label{sec:ourbench}

In this section, we introduce \ourbench, a clinically inspired multi-modal dataset designed to bridge real-world diagnostic workflows and vision–language modeling. We first describe its clinical motivation and guiding design principles, followed by a detailed overview of its multimodal composition and associated reasoning tasks.

\subsection{Overview and Background}

\textit{\ourbench}~is a multimodal gastric cancer dataset specifically curated for vision–language model research. Its design is motivated by how clinicians diagnose gastric cancer in practice. Clinically, diagnosis is never based on a single modality. Radiologists analyze multiple CT phases, gastroenterologists interpret endoscopic findings. These imaging assessments are integrated with biochemical tests and laboratory results to form a comprehensive diagnostic conclusion.

This complex multi-model reliance process inspired the creation of our \textit{\ourbench}: a dataset that mirrors real diagnostic reasoning by aligning these heterogeneous data sources within a unified framework. 
Each patient record in Gastric-X includes quad-phase CT scans, endoscopic images, biochemical indicators, expert bounding box annotations of lesions, detailed imaging and endoscopic reports, and final diagnostic summaries. 
 To facilitate advanced reasoning tasks, Gastric-X is further enriched with  Visual Question Answering (VQA) annotations, making it a comprehensive benchmark for clinical VLM development.  The overview of dataset distribution can be seen in Figure~\ref{fig:stat}.

\subsection{Multi-modal Data}
Our dataset is inherently multi-modal, encompassing diverse types of medical data that reflect how clinicians integrate information from multiple sources during diagnosis.

\smallskip\noindent\textbf{Imaging Modality.} 
The foundation of \textit{\ourbench}~is its diverse and complementary imaging data. We begin with the CT modality, collected across four key phases: arterial, venous, equilibrium, and non-contrast. Each phase offers a unique view of vascular perfusion and tissue enhancement, and together they capture the dynamic passage of contrast through the gastric wall.  In clinical practice, radiologists often integrate these dynamic differences to form a holistic understanding of disease progression. \textit{\ourbench}~reflects this process in data form.

Complementing CT, \textit{\ourbench}~also includes endoscopic images, which provide high-resolution, color-rich views of the gastric mucosa. These images expose fine-grained textures, color variations, and microvascular structures invisible to CT. Subtle irregularities, such as fold distortion, erosion, or discoloration, often signal the earliest stages of malignancy and help pinpoint tumor location and stage.

\textit{\ourbench}~comprises 7.1K CT scans (a total of 83.48K slices) and 1.7K endoscopic images. We unify these complementary views to support a holistic understanding of gastric cancer through multi-modal imaging. The distribution of multi-phase CT slices is shown in Figure~\ref{fig:stat}{\color{cvprblue}C}.


\smallskip\noindent\textbf{Stages and Lesion Annotation.}
Gastric cancer is clinically staged using the TNM system, which characterizes disease progression by assessing the primary tumor (T), regional lymph nodes (N), and distant metastasis (M). This framework guides diagnosis, treatment planning, and prognosis estimation, ultimately determining the overall stage. The distribution of overall stages by gender is shown in Figure~\ref{fig:stat}{\color{cvprblue}A}. The transition from TNM components to overall stages is illustrated in Figure~\ref{fig:stat}{\color{cvprblue}D}, and the tumor size distribution is presented in Figure~\ref{fig:stat}{\color{cvprblue}E}.

For each CT phase, \textit{\ourbench}~provides detailed 3D bounding box (BBox) annotations covering both tumor lesions and relevant organs. The annotations are organized into three hierarchical levels to capture different diagnostic focuses: the tumor core, the regional lymph nodes, and the entire stomach region. In total, each CT study includes three BBoxes per phase across four phases and 1.74K patients, resulting in a rich and comprehensive annotation set of 21,408 BBoxes for multi-scale lesion analysis.

\smallskip\noindent\textbf{Biochemical Data.} 
Beyond imaging, the diagnostic reasoning of clinicians often turns to the language of biochemical test, signals that quietly reflect internal pathology.  
\textit{\ourbench} captures this dimension through {11 serum biochemistry indicators} (\eg~liver and renal function markers) and {5 key tumor markers} commonly used in gastric cancer screening.  
To complement these laboratory measures, we further include comprehensive {electronic health records (EHR)} detailing surgical procedures, medication history, and treatment progress across {134 structured items}.  The tumor marker distribution is shown in  Figure~\ref{fig:stat}{\color{cvprblue}B}.

\smallskip\noindent\textbf{Reports. }
Beyond imaging, \textit{\ourbench}~incorporates rich clinical reports that mirror the reasoning chain of real-world diagnosis. Each patient record is paired with three complementary types of reports: 
(1) the CT report, describing lesion morphology, enhancement patterns, and staging impressions;
(2) the endoscopic report, detailing mucosal appearance, color, and biopsy findings; and
(3) the diagnosis report, which summarizes pathological confirmation and clinical outcomes. The word cloud of diagnosis reports is shown in Figure~\ref{fig:stat}{\color{cvprblue}F}.

Building upon these reports, we construct {26,760 visual question–answer (VQA) pairs}, designed to transform narrative clinical observations into structured reasoning tasks.

\subsection{Comparison to Other Datasets}

As summarized in Table~\ref{tab:dataset_comparison}, existing medical vision–language datasets primarily focus on static imaging modalities (e.g., X-ray, CT, pathology slides) paired with diagnostic reports or VQA annotations. While datasets such as PathVQA~\cite{he2020pathvqa}, SLAKE~\cite{liu2021slake}, and GEMeX~\cite{liu2025gemex} have advanced multimodal research, they lack dynamic information from multi-phase imaging and structured clinical measurements.

More recent datasets, including Merlin~\cite{blankemeier2024merlinrelated3} and MIMIC-CXR~\cite{johnson2024mimiccxr}, begin to incorporate biochemical data. Structured EHR components such as serology results offer quantitative insights into physiological states not captured by imaging alone. 
For gastric cancer specifically, TCGA-STAD~\cite{Lucchesi_Aredes_2016_TCGA_STAD} provides 46 subjects with CT scans and accompanying clinical, genomic, and histopathology data, but remains limited in scale.

In contrast, \textit{\ourbench} introduces a previously missing multimodal configuration: multi-phase 3D CT, endoscopy, structured biochemistry, lesion annotations, and clinical text, all aligned at the patient level. No existing dataset offers such a comprehensive setting. This design provides the multimodal supervision necessary for realistic clinical reasoning and establishes a new standard for holistic medical understanding.

\begin{figure}
    \centering
    \includegraphics[width=1\linewidth]{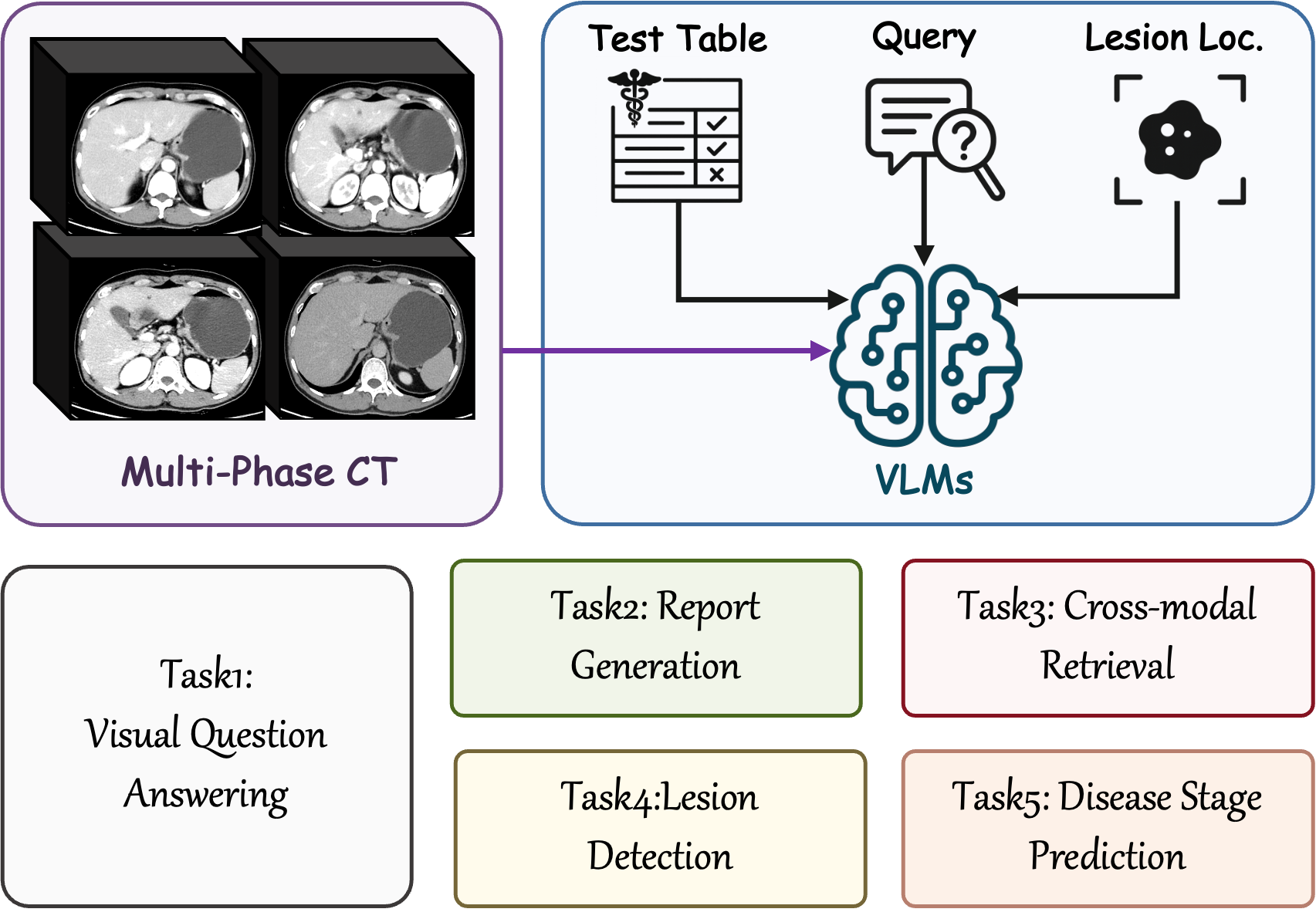}
    \caption{\textbf{VLM adaptation.} In adapting VLMs to our dataset, the visual encoder incorporates multi-phase CT inputs, while test tables, textual queries, and lesion-localization cues serve as complementary multimodal inputs guiding the model’s diagnostic reasoning. The VLMs must effectively adapt to these diverse input modalities to better capture and perform the targeted clinical tasks. }
    \label{fig:adaptation}
\end{figure}

\section{Adapting VLMs to Our Multi-Modal Data}

To evaluate how current vision–language models (VLMs) handle our multi-model medical data, we adapt both general-purpose and medical-specific models to \ourbench.   The symbolic adaptation and the tasks can be seen in Figure~\ref{fig:adaptation}.

\smallskip\noindent\textbf{Model Selection.}
We select LLaVA-1.5-7B~\cite{liu2024improvedLLAVA-1.5}, BLIP-2~\cite{li2023blip2}, and X$^2$-VLM~\cite{zeng2023xvlmmed} as representative general VLMs, and LLaVA-Med v1.5~\cite{li2023llavaMed}, Med-Flamingo~\cite{moor2023medflamingo}, and MedVInT~\cite{zhang2023pmcmedvlnt} as medical-domain models.

\smallskip\noindent\textbf{CT Adaptation.}
Our dataset comprises multi-phase 3D CT volumes, whereas most VLMs are designed for 2D or single-image inputs. For LLaVA-1.5~\cite{liu2024improvedLLAVA-1.5} and BLIP-2~\cite{li2023blip2}, which natively support multi-channel image inputs, we concatenate CT slices across phases to form multi-channel representations. For X$^2$-VLM~\cite{zeng2023xvlmmed}, whose architecture is 2D, we replace its vision encoder with a 3D Swin Transformer~\cite{liu2021swinbase,liu2022video} and its text encoder with MedBERT~\cite{rasmy2021med}, both initialized with pre-trained weights. The adapted model is referred to as X$^2$-VLM-Med. 


For medical-specific models, Med-Flamingo and MedVInT already support volumetric inputs in the form of sequential 2D (pseudo-3D) processing and therefore require only minimal modification. In contrast, LLaVA-Med natively accepts only 2D inputs, so we adopt the same strategy used for X$^2$-VLM~\cite{zeng2023xvlmmed} by replacing its vision encoder with a Swin Transformer~\cite{liu2022video} to enable effective 3D volume processing.

\smallskip\noindent\textbf{Retrieval Extension.}
Some models lack built-in retrieval capability. We add a lightweight retrieval head to enable bidirectional image–text matching, allowing consistent evaluation across all models on cross-modal retrieval tasks.

\smallskip\noindent\textbf{Addition modalities usage.} Clinical diagnosis rarely relies on imaging alone; clinicians naturally integrate visual cues with structured biomedical measurements. To mirror this workflow with minimal architectural changes, we introduce two lightweight but impactful sources of auxiliary information: bounding-box cues and biomedical test tables, to support the VLMs. The bounding boxes are rendered directly on the CT slices as colored overlays, serving as soft spatial priors that nudge the VLM’s attention toward clinically relevant regions without altering the visual encoder.

In contrast, biomedical test tables present a unique challenge: they contain dozens of measurements, many of which are irrelevant or fall within normal ranges. Rather than feeding the entire table, we mimic the behavior of clinicians, who first look for abnormal values, meaning values that exceed physiological thresholds, and reason from there. We extract only these abnormal entries and convert them into concise textual descriptors that include the test name, its measured value, and the ratio by which it exceeds the normal limit.

\section{Experiments}
\label{sec:exp}

\begin{figure}[t]
    \centering
    \includegraphics[width=1\linewidth]{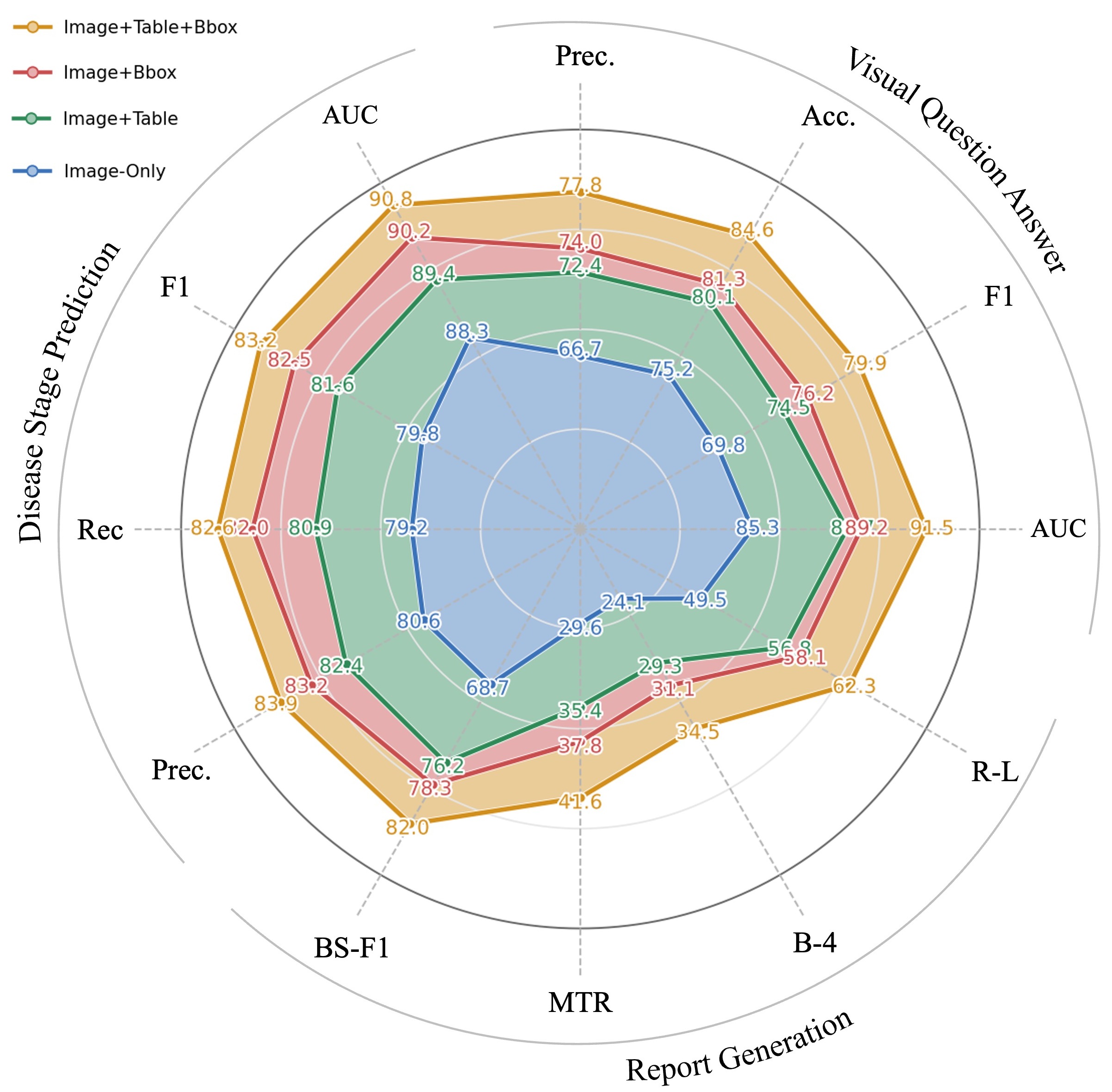}
   \caption{Radar plot comparing multimodal configurations across three medical vision-language tasks. The "Image+Table+Bbox" configuration achieves the highest overall performance across all evaluation metrics.}
    \label{fig:radar_results}
\end{figure}


\textbf{Implementation Details.} All models are fine-tuned using the AdamW optimizer with a batch size of 32 and a learning rate of 5e-5, following a linear warm-up and decay schedule. All models are trained on a single NVIDIA RTX 3090 GPU. The dataset is split at the {patient level into training, validation, and test sets (70/15/15)}.  All models are initialized from the X2-VLM checkpoint and trained independently for each task. We optimize models using {AdamW with a learning rate of 5$\times$10$^{-5}$}, weight decay 0.01, and {10\% warm-up schedule}. The text encoder is trained with a {2$\times$learning rate} relative to the visual encoder.

\newcommand{\mone}{VisionLModel-1}
\newcommand{\mtwo}{VisionLModel-2}
\newcommand{\mthree}{VisionLModel-3}
\newcommand{\mfour}{VisionLModel-4}
\newcommand{\mfive}{VisionLModel-5}


\begin{table*}[t]
\centering
\caption{
Performance across modalities for Vision Question Answer and report generation. 
We evaluate four input-modality settings, including Image Only and combinations that add Table, Bounding Box (BBox), or both. 
The best and second-best results in each column are shown in \best{bold} and \second{underlined}, respectively.
\label{tab:performance_vqa_report}
}

\vspace{-2pt}
\begin{threeparttable}
\setlength{\tabcolsep}{4.5pt}
\subfloat[\textbf{Visual Question Answering.}\label{tab:vqa}  Prec.: Precision, Acc.: Accuracy, AUC: Area Under the ROC Curve.]{
\resizebox{\textwidth}{!}{
\begin{tabular}{l
        S[table-format=2.1, table-number-alignment=center]
        S[table-format=2.1, table-number-alignment=center]
        S[table-format=2.1, table-number-alignment=center]
        S[table-format=2.1, table-number-alignment=center]
        S[table-format=2.1, table-number-alignment=center]
        S[table-format=2.1, table-number-alignment=center]
        S[table-format=2.1, table-number-alignment=center]
        S[table-format=2.1, table-number-alignment=center]
        S[table-format=2.1, table-number-alignment=center]
        S[table-format=2.1, table-number-alignment=center]
        S[table-format=2.1, table-number-alignment=center]
        S[table-format=2.1, table-number-alignment=center]
        S[table-format=2.1, table-number-alignment=center]
        S[table-format=2.1, table-number-alignment=center]
        S[table-format=2.1, table-number-alignment=center]
        S[table-format=2.1, table-number-alignment=center]}
\toprule
\multirow{2}{*}{Model}
& \multicolumn{4}{c}{\io}
& \multicolumn{4}{c}{\itb}
& \multicolumn{4}{c}{\ibb}
& \multicolumn{4}{c}{\itbb} \\
\cmidrule(lr){2-5}\cmidrule(lr){6-9}\cmidrule(lr){10-13}\cmidrule(lr){14-17}
\noalign{\vskip -\arrayrulewidth}
& {Prec. $\uparrow$} & {Acc. $\uparrow$} & {F1 $\uparrow$} & {AUC (\%) $\uparrow$}
& {Prec. $\uparrow$} & {Acc. $\uparrow$} & {F1 $\uparrow$} & {AUC (\%) $\uparrow$}
& {Prec. $\uparrow$} & {Acc. $\uparrow$} & {F1 $\uparrow$} & {AUC (\%) $\uparrow$}
& {Prec. $\uparrow$} & {Acc. $\uparrow$} & {F1 $\uparrow$} & {AUC (\%) $\uparrow$} \\
\midrule
LLaVA-1.5-7B~\cite{liu2024improvedLLAVA-1.5}& 42.5 & 53.1 & 46.2 & 67.8
                & 48.3 & 59.7 & 52.4 & 72.1
                & 50.1 & 61.0 & 54.3 & 73.0
                & 55.2 & 66.5 & 59.1 & 77.8 \\

LLaVA-Med v1.5~\cite{li2023llavaMed}  & 56.3 & 64.7 & 59.9 & 78.4
                & 62.5 & 71.8 & 65.7 & 82.0
                & 63.8 & 72.9 & 66.9 & 82.6
                & 67.4 & 76.1 & 69.8 & 85.0 \\

Med-Flamingo~\cite{moor2023medflamingo}    & \second{60.8} & \second{68.9} & \second{63.9} & \second{80.5}
                & \second{66.9} & \second{74.8} & \second{69.7} & \second{84.2}
                & \second{68.1} & \second{75.4} & \second{70.9} & \second{84.8}
                & \second{71.0} & \second{78.9} & \second{73.5} & \second{86.5} \\

BLIP-2~\cite{li2023blip2}          & 50.2 & 59.4 & 53.7 & 73.1
                & 56.0 & 64.8 & 58.9 & 77.0
                & 58.0 & 66.4 & 60.7 & 78.2
                & 61.3 & 69.7 & 63.9 & 80.1 \\

MedVInT~\cite{zhang2023pmcmedvlnt}         & 58.1 & 66.0 & 60.9 & 79.0
                & 63.4 & 71.5 & 65.9 & 82.5
                & 65.0 & 72.3 & 67.4 & 83.0
                & 68.0 & 75.8 & 70.1 & 85.4 \\

\textbf{X$^2$-VLM-Med}~\cite{zeng2023xvlmmed} 
                & \best{66.7} & \best{75.2} & \best{69.8} & \best{85.3}
                & \best{72.4} & \best{80.1} & \best{74.5} & \best{88.7}
                & \best{74.0} & \best{81.3} & \best{76.2} & \best{89.2}
                & \best{77.8} & \best{84.6} & \best{79.9} & \best{91.5} \\
\bottomrule
\end{tabular}

}
}
\vspace{2mm}
\hfill

\subfloat[\textbf{Report Generation. }\label{tab:report} Evaluation metrics include ROUGE-L (R-L), BLEU-4 (B-4), METEOR (MTR), and BERTScore F1 (BS-F1). ]{
\resizebox{\textwidth}{!}{
\begin{tabular}{l
  S[table-format=2.1] S[table-format=2.1] S[table-format=2.1] S[table-format=2.1]
  S[table-format=2.1] S[table-format=2.1] S[table-format=2.1] S[table-format=2.1]
  S[table-format=2.1] S[table-format=2.1] S[table-format=2.1] S[table-format=2.1]
  S[table-format=2.1] S[table-format=2.1] S[table-format=2.1] S[table-format=2.1]}
\toprule
\multirow{2}{*}{Model}
& \multicolumn{4}{c}{\io}
& \multicolumn{4}{c}{\itb}
& \multicolumn{4}{c}{\ibb}
& \multicolumn{4}{c}{\itbb} \\
\cmidrule(lr){2-5}\cmidrule(lr){6-9}\cmidrule(lr){10-13}\cmidrule(lr){14-17}
& {R-L $\uparrow$} & {B-4 $\uparrow$} & {MTR $\uparrow$} & {BS-F1 $\uparrow$}
& {R-L $\uparrow$} & {B-4 $\uparrow$} & {MTR $\uparrow$} & {BS-F1 $\uparrow$}
& {R-L $\uparrow$} & {B-4 $\uparrow$} & {MTR $\uparrow$} & {BS-F1 $\uparrow$}
& {R-L $\uparrow$} & {B-4 $\uparrow$} & {MTR $\uparrow$} & {BS-F1 $\uparrow$}
 \\
\midrule
LLaVA-1.5-7B~\cite{liu2024improvedLLAVA-1.5}    & 28.4 & 9.2  & 14.1 & 45.3
                & 34.2 & 12.1 & 17.6 & 51.2
                & 36.0 & 13.5 & 18.9 & 53.1
                & 40.1 & 16.3 & 21.7 & 57.8 \\

LLaVA-Med v1.5~\cite{li2023llavaMed}  & 35.7 & 13.8 & 18.9 & 53.6
                & 42.0 & 17.2 & 23.4 & 60.0
                & 44.1 & 18.7 & 25.1 & 62.6
                & 48.7 & 21.9 & 28.4 & 66.5 \\

Med-Flamingo~\cite{moor2023medflamingo}    & \second{42.1} & \second{18.6} & \second{23.4} & \second{60.2}
                & \second{49.0} & \second{22.4} & \second{28.2} & \second{67.0}
                & \second{51.2} & \second{24.1} & \second{30.1} & \second{69.4}
                & \second{55.6} & \second{27.8} & \second{34.6} & \second{73.1} \\

BLIP-2~\cite{li2023blip2}          & 36.8 & 15.2 & 19.7 & 55.1
                & 43.5 & 19.0 & 25.0 & 61.9
                & 45.6 & 20.6 & 26.8 & 64.0
                & 50.2 & 23.7 & 30.5 & 67.7 \\

MedVInT~\cite{zhang2023pmcmedvlnt}         & 38.9 & 16.9 & 21.0 & 57.3
                & 45.1 & 20.8 & 26.9 & 63.5
                & 47.2 & 22.3 & 28.7 & 65.9
                & 52.0 & 25.5 & 32.4 & 69.8 \\

\textbf{X$^2$-VLM-Med}~\cite{zeng2023xvlmmed}
                & \best{49.5} & \best{24.1} & \best{29.6} & \best{68.7}
                & \best{56.8} & \best{29.3} & \best{35.4} & \best{76.2}
                & \best{58.1} & \best{31.1} & \best{37.8} & \best{78.3}
                & \best{62.3} & \best{34.5} & \best{41.6} & \best{82.0} \\

\bottomrule
\end{tabular}
}
}

\end{threeparttable}
\end{table*}

\vspace{2mm}
\noindent \textbf{Multi-Modal Input Settings.} 
In clinical practice, multiple modalities are often examined together to support diagnosis. 
To emulate this process and evaluate the contribution of each modality, we design four input schemas: 
(1) \textit{Image Only}, using visual information alone; 
(2) \textit{Image + Table}, combining images with biomedical test results; 
(3) \textit{Image + BBox}, incorporating both images and bounding box annotations; and 
(4) \textit{Image + Table + BBox}, integrating all available modalities for a comprehensive diagnostic setting.

\vspace{2mm}
\noindent \textbf{Task Settings.} 
To comprehensively evaluate multi-modal understanding, we design five challenges within our dataset: report generation, visual question answering, cross-modal retrieval, phase classification, and lesion detection.

For all task result tables, the best and second-best results are indicated in \best{bold} and \second{underlined}, respectively.

A summary of the model performance across all configurations in the three main tasks is illustrated in Figure~\ref{fig:radar_results}. The results are reported using X$^2$-VLM-Med~\cite{zeng2023xvlmmed}, demonstrating consistent improvements when incorporating multimodal information. Notably, the \textit{Image+Table+Bbox} configuration achieves the most balanced and superior results across all evaluation metrics.

The detailed configurations and results for each task are presented in the following subsections.

\subsection{Visual Question Answering} 

As shown in Table \ref{tab:vqa}, across all four evaluation settings, a consistent upward trend is observed in model performance as the input conditions become progressively richer and complex. Specifically, all models exhibit improvements in Precision, Accuracy, F1, and Area Under the Curve (AUC) from the image-only setting to the fully integrated one, indicating that exposure to more informative or balanced input–target combinations substantially enhances visual–language reasoning.

Among the compared methods, X$^2$-VLM-Med~\cite{zeng2023xvlmmed} achieves the best overall performance across all metrics and settings, demonstrating superior generalization and multimodal alignment capabilities. Its AUC steadily increases from 85.3\% (\io) to 91.5\% (\itbb), reflecting robust discriminative power under varying configurations. Med-Flamingo~\cite{moor2023medflamingo} consistently ranks second, highlighting its strong medical-domain adaptation, while MedVInT~\cite{zhang2023pmcmedvlnt} closely follows with competitive gains across all metrics.

\subsection{Report Generation}

We evaluate report generation quality using four complementary metrics: ROUGE-L (R-L), BLEU-4 (B-4), METEOR (MTR), and BERTScore F1 (BS-F1), to cover lexical overlap, fluency, and semantic fidelity between generated and reference reports.

As shown in Table~\ref{tab:report}, all models exhibit a consistent upward trend across the four experimental settings, mirroring the pattern observed in Table~\ref{tab:vqa}.  X$^2$-VLM-Med~\cite{zeng2023xvlmmed} achieves the best overall performance, with BERTScore F1 improving from 68.7 in \io~to 82.0 in \itbb, highlighting its strong cross-modal reasoning and text generation capabilities. 


\subsection{Cross-modal Retrieval}

Table \ref{tab:retrieval} presents image-only cross-modal retrieval results for both Image→Text and Text→Image, evaluated using Recall at K (Recall@K), Median Rank (MedR), Mean Rank (MnR), and Mean Average Precision (mAP).

Across all metrics, X$^2$-VLM-Med~\cite{zeng2023xvlmmed} achieves the best performance, reaching an R@1 of 48.9\% for Image→Text and 47.5\% for Text→Image, substantially outperforming other models. 
Med-Flamingo~\cite{moor2023medflamingo} consistently ranks second, reflecting effective multimodal reasoning within the medical domain.

MedVInT~\cite{zhang2023pmcmedvlnt} and BLIP-2~\cite{li2023blip2} deliver competitive results but remain behind the top two models, while LLaVA-Med v1.5~\cite{li2023llavaMed} and LLaVA-1.5-7B~\cite{liu2024improvedLLAVA-1.5} lag further due to limited retrieval-specific adaptation.

\subsection{Disease Stage Classification}
Table~\ref{tab:stage} presents the results for disease stage classification across all evaluated models. 
We observe a consistent performance hierarchy across input configurations, broadly aligned with both model capacity and multimodal alignment strength. Conventional convolutional architectures such as ResNet-50~\cite{he2016deep} lag considerably behind transformer-based approaches, reflecting their limited ability to capture long-range and cross-modal dependencies. In contrast, the Swin Transformer~\cite{liu2021swinbase} achieves notable gains over ResNet baselines, confirming that hierarchical self-attention effectively models localized medical features and spatial-scale variations.

Among multimodal encoders, MedVInT~\cite{zhang2023pmcmedvlnt} and LLaVA-Med v1.5~\cite{li2023llavaMed} demonstrate moderate improvements, particularly when incorporating table cues. However, their performance saturates as the modality complexity increases (\itbb).

In contrast, X$^2$-VLM-Med~\cite{zeng2023xvlmmed} consistently achieves the strongest results across all metrics and input configurations, outperforming the second-best model by a clear margin. The gain is most pronounced in AUC (+1.6 over Swin on \itbb).

\begin{table*}[t]
\centering
\caption{\label{tab:retrieval}
\textbf{Cross-modal retrieval.}
This table reports retrieval results in both Image-to-Text and Text-to-Image directions.
Metrics include Recall@K (\%, higher is better), Median Rank and Mean Rank (lower is better), and mean Average Precision (mAP). 
\best{Bold} and \second{underlined} numbers denote the best and second-best performance in each column.
}
\vspace{-2pt}
\setlength{\tabcolsep}{5pt}
\begin{threeparttable}
\resizebox{\textwidth}{!}{
\begin{tabular}{l
  S[table-format=2.1] S[table-format=2.1] S[table-format=2.1]
  S[table-format=2.2] S[table-format=2.1] S[table-format=2.1]
  @{\hskip 15pt}
  S[table-format=2.1] S[table-format=2.1] S[table-format=2.1]
  S[table-format=2.2] S[table-format=2.1] S[table-format=2.1]}
\toprule
\multirow{2}{*}{Model}
& \multicolumn{6}{c}{Image $\rightarrow$ Text} 
& \multicolumn{6}{c}{Text $\rightarrow$ Image} \\
\cmidrule(lr){2-7}\cmidrule(lr){8-13}
& {R@1 $\uparrow$} & {R@5 $\uparrow$} & {R@10 $\uparrow$} 
& {MedR $\downarrow$} & {MnR $\downarrow$} & {mAP $\uparrow$}
& {R@1 $\uparrow$} & {R@5 $\uparrow$} & {R@10 $\uparrow$} 
& {MedR $\downarrow$} & {MnR $\downarrow$} & {mAP $\uparrow$} \\
\midrule
LLaVA-1.5-7B~\cite{liu2024improvedLLAVA-1.5} & 24.3 & 52.7 & 63.1 & 11.2 & 28.4 & 38.5 & 22.1 & 50.3 & 61.7 & 12.4 & 29.8 & 36.9 \\
LLaVA-Med v1.5~\cite{li2023llavaMed} & 35.6 & 68.9 & 78.3 & 7.5 & 19.6 & 52.7 & 33.8 & 65.2 & 76.1 & 8.1 & 21.4 & 50.3 \\
Med-Flamingo~\cite{moor2023medflamingo} & \second{42.8} & \second{74.1} & \second{83.5} & \second{6.2} & \second{17.3} & \second{57.9} & \second{41.5} & \second{72.6} & \second{82.8} & \second{6.6} & \second{18.4} & \second{56.8} \\
BLIP-2~\cite{li2023blip2} & 39.7 & 70.5 & 80.9 & 6.9 & 18.1 & 55.3 & 37.6 & 68.9 & 79.2 & 7.4 & 19.7 & 53.2 \\
MedVInT~\cite{zhang2023pmcmedvlnt} & 40.1 & 73.6 & 82.4 & 6.4 & 17.1 & 56.4 & 39.4 & 71.8 & 81.2 & 6.8 & 17.9 & 55.1 \\
\textbf{X$^2$-VLM-Med}~\cite{zeng2023xvlmmed} & \best{48.9} & \best{80.7} & \best{88.2} & \best{4.9} & \best{13.5} & \best{63.1} & \best{47.5} & \best{79.3} & \best{87.4} & \best{5.2} & \best{14.1} & \best{61.7} \\
\bottomrule
\end{tabular}
}
\end{threeparttable}
\end{table*}

\begin{table*}[t]
\centering
\caption{\label{tab:stage}
\textbf{Disease stage classification.} 
For each configuration, we evaluate Precision, Recall, F1 score, and Area Under (AUC). 
These results show how integrating multimodal cues and medical-aware pretraining benefits fine-grained disease staging.
\best{Bold} and \second{underlined} numbers indicate the best and second-best performance in each column.
}

\vspace{-2pt}
\begin{threeparttable}
\setlength{\tabcolsep}{4.5pt}
\resizebox{\textwidth}{!}{
\begin{tabular}{l
                S[table-format=2.1] S[table-format=2.1] S[table-format=2.1] S[table-format=2.1]
                S[table-format=2.1] S[table-format=2.1] S[table-format=2.1] S[table-format=2.1]
                S[table-format=2.1] S[table-format=2.1] S[table-format=2.1] S[table-format=2.1]
                S[table-format=2.1] S[table-format=2.1] S[table-format=2.1] S[table-format=2.1]}
\toprule
\multirow{2}{*}{\textbf{Model}}
& \multicolumn{4}{c}{\io}
& \multicolumn{4}{c}{\itb}
& \multicolumn{4}{c}{\ibb}
& \multicolumn{4}{c}{\itbb} \\
\cmidrule(lr){2-5}\cmidrule(lr){6-9}\cmidrule(lr){10-13}\cmidrule(lr){14-17}
& {Prec. $\uparrow$} & {Rec. $\uparrow$} & {F1 $\uparrow$} & {AUC $\uparrow$}
& {Prec. $\uparrow$} & {Rec. $\uparrow$} & {F1 $\uparrow$} & {AUC $\uparrow$}
& {Prec. $\uparrow$} & {Rec. $\uparrow$} & {F1 $\uparrow$} & {AUC $\uparrow$}
& {Prec. $\uparrow$} & {Rec. $\uparrow$} & {F1 $\uparrow$} & {AUC $\uparrow$} \\
\midrule
ResNet-50~\cite{he2016deep}              & 72.8 & 70.6 & 71.6 & 80.3 & 74.1 & 72.0 & 73.0 & 81.2 & 75.9 & 73.7 & 74.7 & 82.1 & 76.4 & 74.8 & 75.5 & 82.7 \\
Swin Transformer~\cite{liu2021swinbase}  & \second{79.5} & \second{78.0} & \second{78.7} & \second{87.1}
                       & 80.4 & 79.2 & 79.8 & 88.0 & \second{82.1} & \second{80.9} & \second{81.5} & \second{89.2} & \second{83.4} & \second{82.0} & \second{82.7} & \second{90.1} \\
MedVInT~\cite{zhang2023pmcmedvlnt}         & 77.1 & 76.2 & 76.6 & 85.5 & \second{81.0} & \second{79.8} & \second{80.4} & \second{87.6} & 80.6 & 79.4 & 80.0 & 88.3 & 81.2 & 80.1 & 80.6 & 88.7 \\
LLaVA-Med v1.5~\cite{li2023llavaMed}  & 75.8 & 74.3 & 75.0 & 84.0 & 78.0 & 76.6 & 77.3 & 85.3 & 78.8 & 77.5 & 78.1 & 86.1 & 79.5 & 78.0 & 78.7 & 86.6 \\
\textbf{X$^2$-VLM-Med}~\cite{zeng2023xvlmmed} 
                       & \best{80.6} & \best{79.2} & \best{79.8} & \best{88.3}
                       & \best{82.4} & \best{80.9} & \best{81.6} & \best{89.4}
                       & \best{83.2} & \best{82.0} & \best{82.5} & \best{90.2}
                       & \best{83.9} & \best{82.6} & \best{83.2} & \best{90.8} \\
\bottomrule
\end{tabular}
}
\end{threeparttable}
\end{table*}

\subsection{Lesion Detection}
Table \ref{tab:lesion} summarizes lesion detection performance under COCO evaluation settings \cite{lin2014coco}, reporting COCO-style AP and F1. Faster R-CNN \cite{ren2015fastercnn} serves as a strong convolutional baseline but shows moderate precision, with AP@0.5 = 64.1 and localization accuracy of 70.4.

The Swin Transformer~\cite{wen2020transfer} significantly improves across all metrics, benefiting from hierarchical self-attention that captures multi-scale lesion structures. MedVInT~\cite{zhang2023pmcmedvlnt} further boosts performance, attaining the highest AP@0.5 of 72.1 and a mean AP of 50.2, reflecting superior spatial reasoning through multimodal medical pretraining.

LLaVA-Med v1.5~\cite{liu2024improvedLLAVA-1.5} maintains strong overall detection quality, especially in F1@0.5, suggesting effective vision–language grounding despite limited  adaptation. The X$^2$-VLM-Med~\cite{zeng2023xvlmmed} achieves the best results on most evaluation metrics over baselines by a consistent margin.

\begin{table}[t]
\centering
\caption{\label{tab:lesion}\textbf{Lesion detection}. Metrics are Average Precision (AP) and F1. \textit{mAP} denotes mean AP averaged over IoU thresholds from 0.50 to 0.95 (step 0.05), and localization accuracy (Loc. Acc.) is computed at IoU = 0.5. }
\vspace{-2pt}
\begin{threeparttable}
\setlength{\tabcolsep}{6pt}
\resizebox{0.5\textwidth}{!}{
\begin{tabular}{l
                S[table-format=2.1]
                S[table-format=2.1]
                S[table-format=2.1]
                S[table-format=2.1]
                S[table-format=2.1]}
\toprule
\textbf{Model} 
& {AP@0.5} 
& {AP@0.75} 
& {F1@0.5} 
& {mAP} 
& {Loc. Acc.} \\
\midrule
FRCNN~\cite{ren2015fastercnn}           & 64.1 & 48.7 & 61.0 & 43.2 & 70.4 \\
Swin Transformer~\cite{wen2020transfer}         & 70.5 & 53.9 & 66.7 & 48.9 & 77.1 \\
MedVInT~\cite{zhang2023pmcmedvlnt}                & \textbf{72.1} & \underline{55.1} & 67.8 & \underline{50.2} & \second{78.5} \\
LLaVA-Med v1.5~\cite{liu2024improvedLLAVA-1.5}         & 68.3 & 51.7 & \second{67.9} & 46.8 & 75.2 \\
\textbf{X$^2$-VLM-Med}~\cite{zeng2023xvlmmed}     & \second{70.4} & \best{56.4} & \best{68.1} & \best{51.5} & \best{79.6} \\
\bottomrule
\end{tabular}
}
\end{threeparttable}
\vspace{-1em}
\flushleft{\footnotesize 
Metric@IoU indicates computation at a given IoU threshold. 
}
\vspace{-2em}
\end{table}

\section{Ethical Concern and Publication}
\label{sec:ethical}

All data were collected by first-line clinicians under strict institutional ethical approval and in full compliance with relevant privacy and data-protection regulations. After collection, all samples were fully de-identified and manually checked to ensure that no personally identifiable information remained.

The annotations were created by experienced clinicians with domain expertise and were cross-checked for consistency. We aim to ensure that the clinicians' qualifications and the review process provide a solid basis for the reliability and clinical validity of the annotations, which may benefit future research and model development. 

The dataset is well-structured, has obtained Institutional Review Board (IRB) approval, and is planned to be made publicly available after the paper is accepted. The release will follow institutional data-sharing policies, and no data that restrict public or research redistribution will be included. We plan to host a small subset of samples on Hugging Face for demonstration purposes, while the complete dataset will be distributed through our project webpage after publication. Access to the dataset will require signing a consent form prior to distribution. The dataset is intended to be released under the CC BY-NC-ND 4.0 license.

\section{Conclusion}
\label{sec:conclusion}

We presented \textbf{Gastric-X}, a comprehensive multimodal benchmark designed to advance vision–language research in gastric cancer analysis. The dataset is carefully curated from real clinical workflows, integrating multi-phase 3D CT scans, endoscopic images, biochemical indicators, and clinical reports with bounding box and disease stage annotations. Gastric-X provides a unified platform encompassing five core tasks, including visual question answering, report generation, cross-modal retrieval, disease stage classification, and lesion detection, offering a holistic evaluation of multimodal reasoning in medical AI. We will publicly release the dataset and accompanying experimental code to support reproducibility and community development. We envision Gastric-X as a standard reference for developing robust and clinically aligned vision–language models in healthcare.

{
    \small
    \bibliographystyle{ieeenat_fullname}
    \bibliography{main}
}


\clearpage

\appendix

{\noindent \LARGE \bfseries Appendix}


\section{Abstract}

This supplementary document provides extended technical details and additional discussions for the Gastric-X benchmark. Specifically, we present:
(1) a description of the multi-phase CT normalization and alignment pipeline in Sec.~\ref{sec:MCT};
(2) detailed explanations of the provided clinical reports (CT imaging descriptions, endoscopy reports, and diagnostic conclusions) in Sec.~\ref{sec:reports};
(3) a description of the creation, verification, and prompting strategy for all VQA pairs in Sec.~\ref{sec:vqa};
(4) an illustration of the 134 biomedical indicators in Sec.~\ref{sec:indicator}.


\section{Multi-phase CT Standardization Details}
\label{sec:MCT}

Multi-phase CT scans encompass substantial variation across patients and acquisition phases. Our preprocessing pipeline aims to harmonize  patterns, standardize geometric properties, and ensure spatial alignment across phases. 

\smallskip\noindent\textbf{Intensity normalization across phases.}
We apply a unified clipping window of \([-100, 300]\) HU, consistent with recommended gastric soft-tissue imaging ranges. For each CT volume, per-volume z-score normalization is performed after clipping. In addition, we use histogram matching across phases to reduce heterogeneity.



\smallskip\noindent\textbf{Voxel spacing standardization.}
All phases are resampled to isotropic spacing of \(1.0 \times 1.0 \times 1.0\,\text{mm}^3\) using trilinear interpolation for image data.

\smallskip\noindent\textbf{Handling different-sized CT slices.}
Raw scans contain variable numbers of axial slices. Each patient is associated with a coarse 3D bounding region around the stomach, manually annotated by clinical readers. These bounding boxes vary between  
\[
256\times256\times160 \text{ and } 288\times288\times192
\]
depending on patient-specific anatomy. Volumes are cropped or padded to the unified shape: $288\times288\times192$.

\smallskip\noindent\textbf{Multi-phase alignment.}
Arterial and delayed phases are rigidly registered to the venous phase using a 6-DOF transformation optimized via mutual information. Registration is implemented with SimpleITK (Elastix backend). 

\smallskip\noindent\textbf{Quality control.}
Scans with corrupted slices, missing metadata, or excessive misalignment are excluded. Approximately 3–4\% of volumes are filtered by this process.

\section{The Clinical Reports}
\label{sec:reports}
Each patient record contains three types of clinical reports:

\smallskip\noindent\textbf{CT imaging description report.}
A detailed morphological description authored by radiologists. It covers wall thickening, ulceration, enhancement patterns, perigastric fat infiltration, lymph node size/morphology, and incidental findings.

\smallskip\noindent\textbf{Endoscopy report.}
This endoscopy report provides a detailed assessment of the gastrointestinal mucosa, including evaluation of surface texture, ulceration, pit-pattern characteristics, and any other notable structural changes. Lesions are described with explicit documentation of their location, extent, and depth. Biopsy samples, when obtained, are recorded with corresponding anatomical sites to support accurate histologic correlation.



\smallskip\noindent\textbf{Diagnostic conclusion report.}
A concise interpretive summary presenting the radiologist’s overall impression, including features suggestive of malignancy, estimated TNM staging when applicable, assessment of regional or distant nodal involvement, and any pertinent recommendations for further evaluation or correlation with clinical or pathological findings.



\section{The Creation of VQA Pairs}
\label{sec:vqa}
The dataset contains 26,760 VQA pairs derived from clinical  reports. Their creation follows a multi-stage pipeline designed to ensure clinical correctness, semantic grounding, and diversity of reasoning patterns.

\smallskip\noindent\textbf{(1) Large-scale candidate generation using multiple LLMs.}
Two publicly available large language models, e.g., ChatGPT 4.0, Gemini 2.5 and Claude Sonnet 4.0 were prompted with structured instructions to generate initial question candidates.  
The prompts targeted clinically meaningful aspects such as lesion characterization, enhancement behavior across phases, staging-relevant findings, anatomical localization, and factual consistency checks. 
Among all generated candidates, ChatGPT 4.0 contributed 78.32\% of the questions that ultimately passed clinical verification. 

\smallskip\noindent\textbf{(2) Prompt design and controlled extraction.}
To systematically guide generation, we defined five prompt categories:  
(1) lesion-centric question design,  
(2) enhancement-phase reasoning,  
(3) staging-related reasoning,  
(4) anatomical localization questions, and  
(5) binary Yes/No factual verification.  
Each prompt type was designed to reflect reasoning processes typically employed in abdominal radiology. 
Outputs containing ambiguous phrasing or information not present in the source report were automatically removed. 

\begin{table}[t]
\caption{Effectiveness of different prompting strategies for generating clinically valid VQA questions. Validity represents the percentage of Q/A pairs confirmed by both clinicians.}
\centering
\small
\setlength{\tabcolsep}{4pt} 
\begin{tabular}{p{2.7cm} p{1.6cm} p{3.5cm}}
\toprule
\textbf{Prompt Type} & \textbf{Validity (\%)} & \textbf{Remarks} \\
\midrule
Lesion-focused      & 92.4 & Most clinically reliable and consistently grounded. \\
Staging-focused     & 88.1 & Dependent on level of staging detail documented. \\
Enhancement-phase   & 84.7 & Sensitive to phase contrast variations. \\
Localization        & 79.3 & Occasional ambiguity in spatial descriptions. \\
Yes/No factual      & 90.5 & High factual precision but limited question diversity. \\
\bottomrule
\end{tabular}
\label{tab:prompt_effect}
\end{table}

\smallskip\noindent\textbf{(3) Final VQA selection and answer fidelity.}
Only Q/A pairs that strictly adhered to source-report evidence were retained. 
All answers are derived exclusively from the original CT imaging descriptions or diagnostic conclusions, without augmentation using external medical knowledge. 


\smallskip\noindent\textbf{(4) Double-blind clinical verification.}
All candidate Q/A pairs underwent sentence-level verification by two independent clinical experts: a radiologist with seven years of experience and a gastroenterology specialist with ten years of experience. 
Each clinician evaluated the factual correctness of both the question and its corresponding answer by directly comparing them to the source report. 
Discrepancies were flagged and resolved through consensus. 
This process ensures that the final VQA set reflects clinically valid reasoning and avoids hallucinated associations.

\smallskip\noindent\textbf{Prompt effectiveness comparison.}
We summarize the effectiveness of each prompt category in Table~\ref{tab:prompt_effect}, demonstrating that lesion-focused prompts yield the highest clinical validity, while localization prompts exhibit slightly lower consistency due to occasional ambiguity in spatial references.

\section{The Biomedical Indicators}
\label{sec:indicator}
The dataset includes 134 structured biomedical indicators encompassing demographic data, laboratory tests, tumor biomarkers, imaging metadata, surgical information, pathological staging, histological findings, and postoperative outcomes are shown in Table~\ref{tab:full_biomedical_table}. These indicators originate from structured EHR  and were processed to ensure consistency across patients.

All sensitive identifiers (including patient names, ID numbers, phone numbers, and hospitalization codes) were removed or replaced with anonymized pseudonyms. 
Variables unrelated to model training (such as historical comorbidities or unused surgery-related entries), are retained for completeness but marked as not used in this study.





\clearpage
\onecolumn
\small
\setlength{\tabcolsep}{3pt}
\renewcommand{\arraystretch}{1.0}
\small

{
\footnotesize   
\begin{longtable}{p{0.22\linewidth} p{0.28\linewidth} p{0.22\linewidth} p{0.23\linewidth}}
\caption{Full List of 134 Structured Biomedical Variables in the Gastric-X Dataset. De-identified is marked as "De-ID".}
\label{tab:full_biomedical_table} \\
\toprule
Item & Description & Item & Description \\
\midrule
\endfirsthead

\toprule
Item & Description & Item & Description \\
\midrule
\endhead

\midrule
\multicolumn{4}{r}{Continued on next page} \\
\endfoot

\bottomrule
\endlastfoot

\multicolumn{4}{l}{\textbf{Demographics}} \\
Hospital ID (De-ID) & Anonymized hospitalization identifier &
Patient Name (De-ID) & Anonymized patient code \\
Sex & Biological sex &
Age (De-ID) & Age at admission \\
Bed Number (De-ID) & Anonymized bed assignment &
Surgery Date (De-ID) & Date of surgery (De-ID) \\
Imaging ID & CT imaging identifier &
CT Description & Radiology description \\

\midrule
\multicolumn{4}{l}{\textbf{CBC}} \\
CBC Test Date & Date of CBC test &
CBC White Blood Cell Count & White blood cell count \\
CBC Neutrophil Count & Absolute neutrophil count &
CBC Neutrophil Ratio & Neutrophil percentage \\
CBC Lymphocyte Count & Absolute lymphocyte count &
CBC Lymphocyte Ratio & Lymphocyte percentage \\
CBC Hemoglobin & Hemoglobin concentration &
CBC Platelet Count & Platelet count \\

\midrule
\multicolumn{4}{l}{\textbf{Biochemistry}} \\
Biochemistry Test Date & Date of biochemistry test &
Biochemistry Fasting Glucose & Fasting plasma glucose \\
Biochemistry Prealbumin & Serum prealbumin &
Biochemistry ALT & Alanine aminotransferase \\
Biochemistry AST & Aspartate aminotransferase &
Biochemistry Total Protein & Total serum protein \\
Biochemistry Albumin & Serum albumin &
Biochemistry Total Bilirubin & Total bilirubin \\
Biochemistry Direct Bilirubin & Direct bilirubin &
Biochemistry Creatinine & Serum creatinine \\

\midrule
\multicolumn{4}{l}{\textbf{Tumor Markers}} \\
Biochemistry Urea (BUN) & Blood urea nitrogen &
Tumor Markers Test Date & Date of tumor marker test \\
Tumor Markers AFP & Alpha-fetoprotein &
Tumor Markers CEA & Carcinoembryonic antigen \\
Tumor Markers CA125 & Cancer antigen 125 &
Tumor Markers CA724 & Cancer antigen 724 \\
Tumor Markers CA199 & Cancer antigen 19-9 &
Past Medical History  & Past conditions (not used) \\

\midrule
\multicolumn{4}{l}{\textbf{Surgery Details}} \\
Surgery Date & Date of surgery &
Resection Range & Extent of resection \\
Gastrointestinal Reconstruction & Postoperative reconstruction type &
Occupation & Patient occupation \\
Education Level & Highest educational level &
Marital Status & Marital status \\
Ethnicity & Ethnic group &
Admission Method & Mode of admission \\
Insurance Status & Insurance coverage &
ID Number (De-ID) & Anonymized ID number \\

\midrule
\multicolumn{4}{l}{\textbf{Surgical and Admission Info}} \\
Contact Number (De-ID) & Contact phone &
Surgery Admission Date (De-ID) & Admission date for surgery \\
Surgery Discharge Date (De-ID) & Discharge date &
Surgery Hospitalization Cost & Total hospital cost \\
Admission Temperature & Temperature at admission &
Admission Pulse & Pulse rate at admission \\
Admission Respiration & Respiratory rate &
Admission Systolic Pressure & Systolic BP \\
Admission Diastolic Pressure & Diastolic BP &
Height & Height \\
Weight & Weight &
BMI & Body mass index \\
General Condition & Performance status &
Weight Loss & Recent weight loss \\
Reduced Food Intake & Reduced oral intake &
Smoking Status & Smoking history \\
Drinking Status & Alcohol use &
Endoscopy Date (De-ID) & Endoscopy date \\
Endoscopy Tumor Location & Tumor location &
Endoscopy Tumor Size & Tumor size \\
Endoscopy Gross Type & Gross morphology &
Endoscopy Biopsy Pathology & Biopsy pathology \\
Endoscopy Appearance & Visual findings &
Chief Surgeon (De-ID) & Operating surgeon \\

\midrule
\multicolumn{4}{l}{\textbf{Tumor Anatomy and Pathology}} \\
Tumor Anatomical Location & Tumor site &
Maximum Tumor Diameter & Maximal diameter \\
Serosal Invasion & Serosal involvement &
Gross Tumor Type & Macroscopic type \\
Linitis Plastica & Linitis plastica presence &
Perigastric Lymph Nodes & Perigastric node status \\
Liver Metastasis & Liver metastasis &
Adjacent Organ Invasion & Neighboring organ invasion \\
Peritoneal Seeding & Peritoneal metastasis &
Ascites & Ascites presence \\
Pathology ID (De-ID) & Specimen ID &
Tumor Size (Long Axis) & Long axis size \\
Tumor Size (Short Axis) & Short axis size &
Tumor Size (Height) & Height \\
Histologic Grade & Differentiation grade &
Additional Histologic Type & Additional components \\
Distance to Proximal Margin & Proximal margin distance &
Distance to Distal Margin & Distal margin distance \\
Specimen Proximal Margin & Proximal margin status &
Specimen Distal Margin & Distal margin status \\
Perineural Invasion & Perineural invasion &
Vascular Cancer Thrombus & Vascular invasion \\

\midrule
\multicolumn{4}{l}{\textbf{Staging and Lymph Nodes}} \\
T Stage & Pathologic T category &
N Stage & Pathologic N category \\
M Stage & Pathologic M category &
Overall Stage & Final stage \\
Positive Lymph Nodes & Count of positive nodes &
Dissected Lymph Nodes & Total dissected nodes \\

\midrule
\multicolumn{4}{l}{\textbf{Molecular and IHC}} \\
Ki-67 & Proliferation index &
HER2 Status & HER2 expression \\
CLDN Status & Claudin status &
MLH1 & MLH1 expression \\
PMS2 & PMS2 expression &
MSH2 & MSH2 expression \\
MSH6 & MSH6 expression &
EBER & EBV RNA (ISH) \\
PD-L1 Score & PD-L1 score &
MMR Status & Mismatch repair status \\

\midrule
\multicolumn{4}{l}{\textbf{Complications and Outcomes}} \\
Complication Severity & Clavien–Dindo grade &
Secondary Surgery & Reoperation \\
Complication Occurrence & Any complication &
Severe Complication Occurrence & Severe complication \\
Total Hospital Stay & Total days &
Postoperative Stay & Days after surgery \\
Postoperative Fever & Fever occurrence &
Fever Days & Number of fever days \\
Complication Category & Type of complication &
Intervention & Treatment measures \\
Complication Notes & Additional notes & \\

\end{longtable}

}

\end{document}